\documentclass[letterpaper, 10 pt, conference]{ieeeconf}

\makeatletter
\let\NAT@parse\undefined
\makeatother

\IEEEoverridecommandlockouts                              

\usepackage{graphics} 
\usepackage{epsfig} 
\usepackage{mathptmx} 
\usepackage{amsmath} 
\usepackage{amssymb}  
\usepackage{hyperref}
\usepackage[numbers,sort&compress]{natbib}

\usepackage[nolist,nohyperlinks]{acronym}

\title{\LARGE \bf
Go Fetch: Mobile Manipulation in Unstructured Environments
}

\author{Kenneth Blomqvist$^{*}$, Michel Breyer$^{*}$, Andrei Cramariuc$^{*}$, Julian F\"{o}rster$^{*}$, Margarita Grinvald$^{*}$, \\Florian Tschopp$^{*}$, Jen Jen Chung, Lionel Ott$^{1}$, Juan Nieto, Roland Siegwart 
\thanks{$^{*}$ Equal contribution}
\thanks{$^{1}$ School of Information Technologies, University of Sydney, Australia.}
\thanks{All other authors are with the Autonomous Systems Lab, ETH Zürich, Switzerland.}
\thanks{This work was financially supported in part by the Max Planck ETH Center for Learning Systems, the Luxembourg National Research Fund (FNR) 12571953, and ABB Corporate Research.}
}

\begin{document}

\maketitle
\thispagestyle{empty}
\pagestyle{empty}


\begin{abstract}
With humankind facing new and increasingly large-scale challenges in the medical and domestic spheres, automation of the service sector carries a tremendous potential for improved efficiency, quality, and safety of operations.
Mobile robotics can offer solutions with a high degree of mobility and dexterity, however these complex systems require a multitude of heterogeneous components to be carefully integrated into one consistent framework.
This work presents a mobile manipulation system that combines perception, localization, navigation, motion planning and grasping skills into one common workflow for fetch and carry applications in unstructured indoor environments.
The tight integration across the various modules is experimentally demonstrated on the task of finding a commonly-available object in an office environment, grasping it, and delivering it to a desired drop-off location. The accompanying video is available at  \url{https://youtu.be/e89_Xg1sLnY}.
\end{abstract}

\section{INTRODUCTION}

In order to perform tasks in a human environment that has not been specifically designed for robotic purposes, a robot needs to be able to perceive its surroundings, plan motions to avoid unwanted collisions, dexterously manipulate objects and use tools to achieve its goals. Yet, perception is hard due to limitations in both the sensor hardware as well as the technology needed to process the collected data. Even in the best possible cases, only a fraction of the information about the environment can be sensed at any given point in time. Information has to be aggregated into some internal representation as the robot goes about its business. Using this imperfect information, the robot must plan and execute actions while avoiding possible dangers, such as collisions.

To explore the limits of mobile manipulation systems today, we developed a robotic system combining state-of-the-art approaches for mapping, localization, navigation and handling manipulation tasks. Our robot, RoyalYumi, is an ABB YuMi robot mounted on top of an omnidirectional mobile base. It can localize itself within its environment, find, grasp and retrieve common household objects.

In this paper, we present our system and demonstrate its performance on an object retrieval task where the robot has to find, pick up and bring an object to a specified location. We identify several shortcomings in our system and its sub-modules. We highlight areas that are ripe for being addressed by future work on mobile manipulation systems.

\begin{figure}[!t]
\centering
\includegraphics[width=0.75\linewidth]{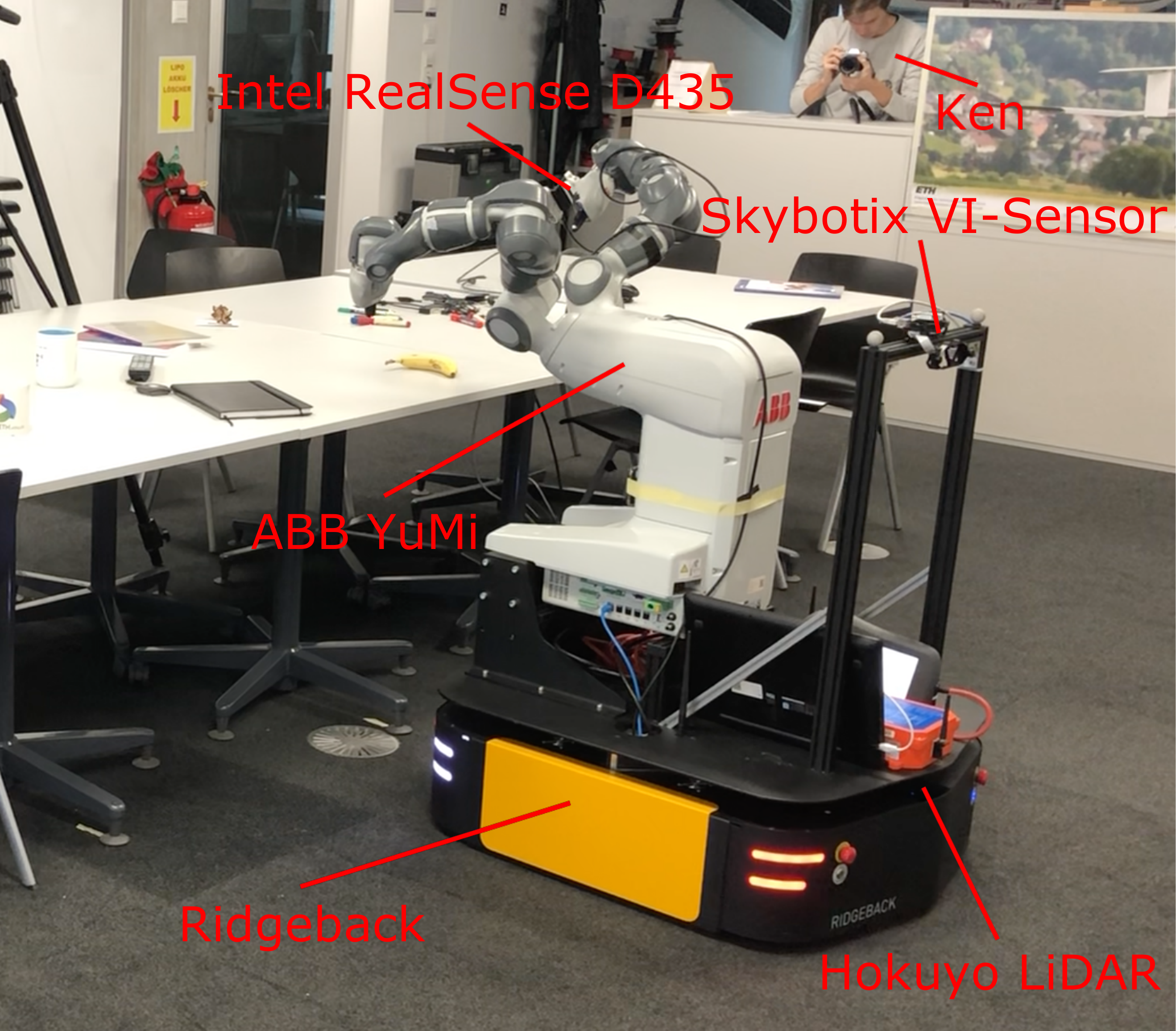}
\caption{A picture of RoyalYumi in action. It features a two-arm ABB Yumi, a Clearpath Ridgeback mobile base, two Hokuyo 2D LiDARs, an Intel RealSense D435 and a Skybotix VI-Sensor.}
\end{figure}

\section{RELATED WORK}
A lot of ink has been spilled on manipulation systems, mobile or otherwise. We highlight the most closely related systems-level papers.

One of the earliest examples of a mobile manipulator is HERB~\cite{srinivasa2010herb}, which was able to search for and retrieve objects in household environments. At the time, the state of the art was such that the system required checkerboard installations in areas demanding high localization accuracy, while action execution was largely open-loop due to the heavy computational cost of each algorithmic component of the system. Nevertheless, several major paradigms established in this seminal work have continued to inform the design and development of robotic mobile manipulators.

Perhaps most prominent among these is the separation of manipulation and navigation tasks. In a typical mobile manipulation scenario, the robot either navigates to a given location~\cite{jain2010assistive,bohren2011towards} or conducts a search routine through the environment to move the platform within reach of the target object~\cite{hollinger2009combining,Bagnell2012,stuckler2016mobile}. Object detection is carried out using a perception pipeline that is distinct from that used for collision avoidance or localization. This means there are two sets of sensors onboard the platform, one for navigation (usually using LiDAR), and another for object detection and interaction (typically RGB-D). While some works perform multi-sensor fusion to improve the navigation map~\cite{Bagnell2012}, these systems still maintain two separate environment representations for manipulation tasks and navigation tasks.

The distinction between these two components is further driven by the behavior-based high-level planning methods popular to mobile manipulation scenarios. Behavior-based planning routines operate on higher-level skills such as ``move to point’’ or ``scan object’’. This allows for planning over long execution sequences, such as those required in the Robocup@Home competitions~\cite{stuckler2016mobile,memmesheimer2017homer}, without incurring the computational overhead needed to plan for individual motor commands. Implementations vary from using simple state machines for switching between behaviors~\cite{jain2010assistive}, to hierarchical state machines, which are capable of more complex task representations~\cite{bohren2011towards}, or the direct use of behavior trees~\cite{Bagnell2012}.

Of course, when planning at such an abstracted layer, failure detection and correction can become very challenging. More recent works such as \cite{memmesheimer2017homer} attempt to improve system robustness through the use of deep learning methods, which have demonstrated strong performance particularly for object recognition~\cite{lin2014microsoft,he2017mask} and grasp detection tasks~\cite{mahler2017dexnet,gualtieri2016high}. While others have also investigated the benefits of learning from human demonstrations~\cite{bajracharya2019mobile}.

\section{SYSTEM}

In our work, we integrate our latest contributions in 3D perception~\cite{grinvald2019volumetric}, mapping and localization~\cite{schneider2018maplab} with state-of-the-art object detection~\cite{redmon2018yolov3} and grasp planning~\cite{tenpas2017grasp} to enable our mobile manipulation platform RoyalYumi to perform an autonomous fetching task.

\subsection{Hardware Setup}

We mounted a two-armed ABB YuMi manipulator on top of an omnidirectional Clearpath Ridgeback mobile base. The Ridgeback is equipped with two Hokuyo 2D LiDAR sensors, both front and back with a 270 degree field of view each. An Intel RealSense D435 RGB-D sensor is mounted on the right wrist of the robot for dense sensing of the workspace and object detection. Finally, a visual-inertial sensor~\cite{nikolic2014synchronized} is mounted on the back of the Ridgeback. The visual-inertial sensor is used to perform state estimation and localization (see Section~\ref{sec:mapping}). It has two monochrome global shutter cameras tightly synchronized with an IMU. One camera is pointed backwards and the other camera is pointed at the ceiling to make sure there are always features to track.

\subsection{Mapping and Navigation} \label{sec:mapping}
In order to perform the first task of finding the target object, RoyalYumi must navigate autonomously in the search environment. 
For this purpose, a globally consistent sparse 3D map of the whole environment is built beforehand using maplab~\cite{schneider2018maplab}. Furthermore, the 2D LiDAR measurements are accumulated using maplab poses to build an occupancy grid used for navigation.

\begin{figure}
    \centering
    \includegraphics[width=\columnwidth]{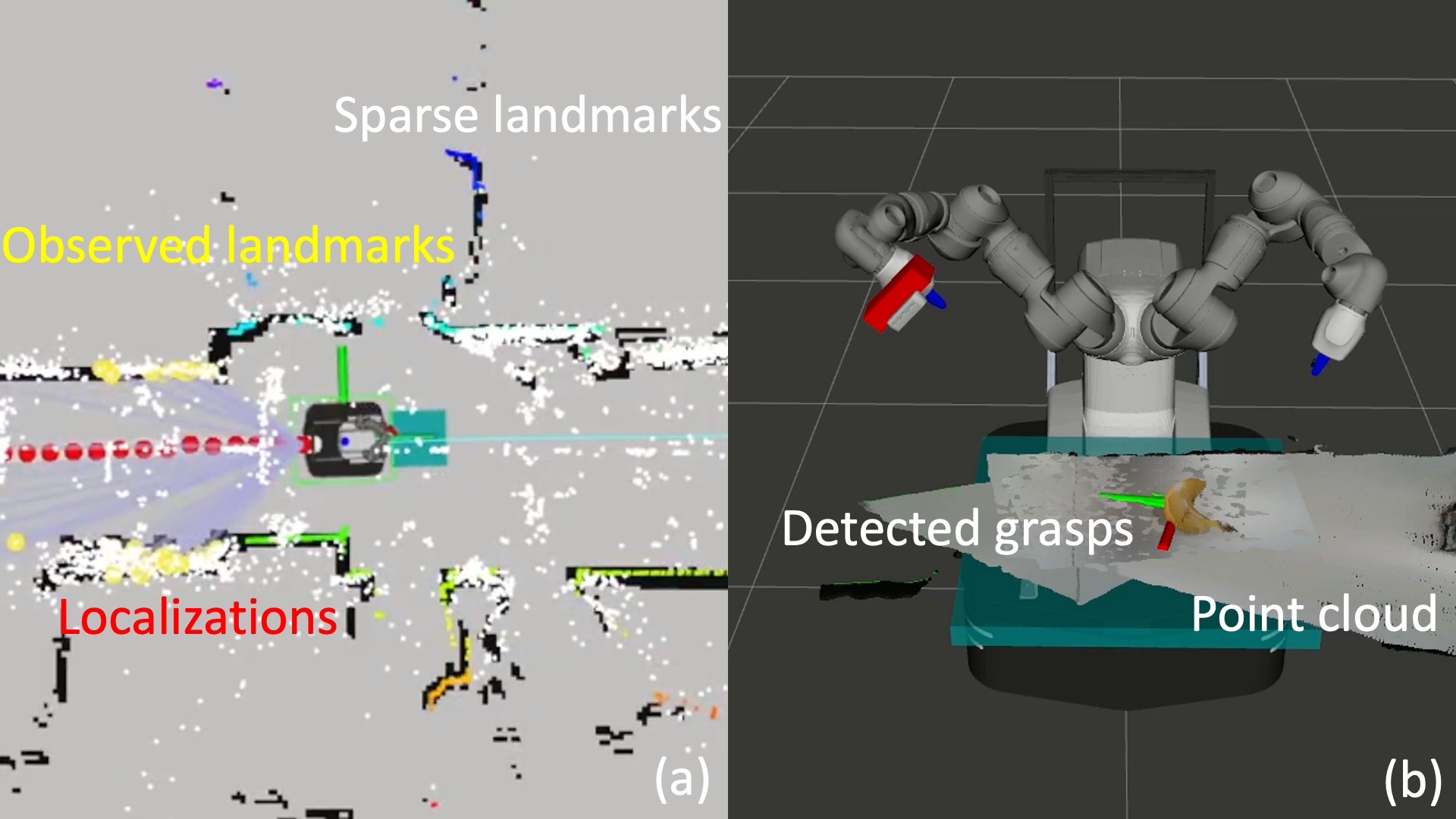}
    \caption{(a) Global localization in a sparse keypoint map using maplab~\cite{schneider2018maplab}. (b) Point cloud reconstruction of the scene and grasp candidates generated by Grasp Pose Detection~\cite{gualtieri2016high}.}
    \label{fig:rviz}
\end{figure}

During online operation, ROVIOLI~\cite{schneider2018maplab} performs online state estimation and localization in the global map based on the sparse features shown in Figure~\ref{fig:rviz} (a). This global position in the map is then used together with the occupancy grid and fixed waypoints (see Section~\ref{sec:high-level}) to perform global navigation by leveraging the capabilities of the ROS navigation stack\footnote{\url{http://wiki.ros.org/move_base}}. Furthermore, ROS navigation is also used to perform local navigation and obstacle avoidance.

\subsection{Object Detection and Object Mapping}
During navigation, the system is continuously trying to find the target object using the RealSense's RGB camera.

For object detection we use the YOLOv3~\cite{redmon2018yolov3} network architecture pre-trained on the COCO dataset~\cite{lin2014microsoft}. The network is run at 2\,Hz, and we only consider those detections whose semantic class corresponds to that of the target object. For each valid detection we initialize a separate image-based tracker~\cite{henriques2014high}, that runs at the same frequency as the camera and propagates the bounding box across frames. Tracked bounding boxes are associated with new detections based on a sufficiently large intersection over union.

Once an object has been tracked across at least 30 frames, we project the center of the detected bounding boxes into 3D space based on the camera extrinsic parameters. The projections together with the corresponding camera poses obtained from the mapping module are then used to triangulate the position of the object. This position is sent to the \textit{approach} module (described in Section \ref{sec:high-level}), which positions the base of the robot as close as possible and facing the object.

\subsection{Dense Reconstruction}
\label{sec:reconstruction}
Besides computing the position of the target object in the environment, the presented approach includes a step for reconstructing the dense 3D geometry of the scene when in proximity to the object of interest.
To this end, the stream of the wrist-mounted RGB-D camera is processed in real-time with the Voxblox++~\cite{grinvald2019volumetric} framework which incrementally builds a volumetric object-level semantic map of the scene.
Such a map provides information about the \mbox{6-\ac{dof}} pose of the target object in the environment and densely describes its 3D shape through a \ac{tsdf} grid.
Further, in the presence of multiple objects in the scene, it can
disambiguate across several instances and identify the desired target object by its semantic class. 

For the captured 3D model of the object of interest to be as complete as possible, during scanning the manipulator arm that is equipped with the RGB-D sensor is programmed to follow a trajectory consisting of four predefined views around the target object.
Once the scanning trajectory is complete, the resulting volumetric reconstruction is converted to a point cloud by extracting the vertices of the surface mesh computed at the zero crossings of the \ac{tsdf} grid.

It is worth noting that the semantic mapping in Voxblox++~\cite{grinvald2019volumetric} relies on the use of a GPU for real-time object detection and segmentation with Mask R-CNN~\cite{he2017mask}.
Should such hardware not be available, a simplified fallback strategy for densely reconstructing the scene of interest would be to execute the scanning trajectory with the manipulator arm and directly stitch together the point clouds captured by the RGB-D camera at each of the four views.
Such an approach, however, cannot unambiguously identify the target element should multiple objects populate the scene of interest.

\subsection{Grasp Planning}

Recent works in grasp planning have moved towards deep learning approaches to directly detect grasps in sensor data. In this work, we use \ac{gpd}~\cite{gualtieri2016high} to generate and classify a large set of 6-\ac{dof} grasp candidates. The algorithm takes as input the point cloud representation of the scene extracted from the Voxblox++ \ac{tsdf} and returns a list of scored grasp hypotheses. We then use MoveIt~\cite{chitta2012moveit} and a ROS interface to YuMi's onboard controllers\footnote{\url{https://github.com/kth-ros-pkg/yumi}} to plan and execute a feasible trajectory to the highest ranked grasp pose, close the gripper, and retrieve the object from the table. 

\subsection{High-level Planning and Execution} \label{sec:high-level}

To simplify the definition of tasks, we designed high-level actions based on the components described above.

A \textit{search action} uses the navigation stack to follow a set of waypoints while running the object detection. It stops successfully once the object is detected and localized in the map.
The \textit{approach action} then computes and navigates the robot to a collision-free pose, positioning the robot as close as possible to the target object while facing it.

Once positioned, the \textit{scan action} moves the wrist-mounted camera above the table top scene for dense reconstruction, feeds the resulting point cloud to the grasp planner and selects the best grasp. If the selected grasp is out of reach of the robot arm, i.e. reaching the grasp pose is kinematically infeasible, this action can also re-position the base provided the object was observed within the initial scans.

Once an in-reach grasp pose is found, the grasp can be executed using the \textit{grasp action}.
Finally, the \textit{drop action} navigates the platform to a user-defined drop location and moves the arm holding the target object to drop it.


All high-level actions of our system are implemented as ROS action servers\footnote{\url{http://wiki.ros.org/actionlib}}. A state machine with a pre-defined action sequence and transition conditions is then used to call and monitor these actions at runtime. The unified interface to our actions simplified and accelerated the process of designing robot behaviors.

\section{EXPERIMENTS}

\begin{figure}
\includegraphics[width=0.95\linewidth]{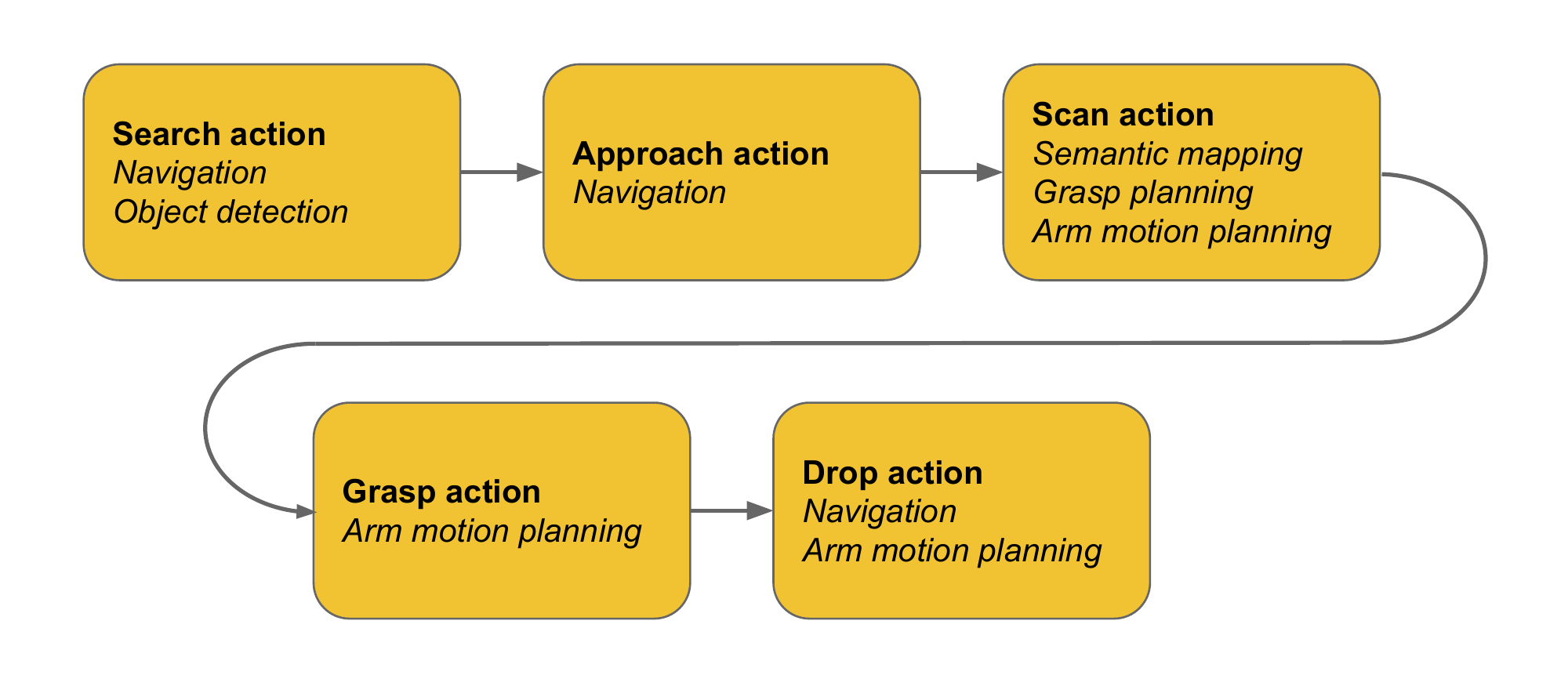}
\centering
\caption{Sequence of high-level actions (Section \ref{sec:high-level}) in the example task that is used to showcase the proposed system. Modules that the respective action relies on are stated in \textit{italic}.}
\label{fig:actionsequence}
\end{figure}

We tested our system on a long-horizon mobile manipulation task. The goal of the task is to find and retrieve an object of interest in our lab. We chose a banana as the target object. The sequence of high-level actions to be used by the state machine was tailored to this task. The actions are illustrated in Figure~\ref{fig:actionsequence}.
Note that in the experimental setup the target object is lying flat on a table and with no other graspable objects nearby.
For this reason, as described in Section~\ref{sec:reconstruction}, the dense 3D reconstruction of the object of interest was done using the simplified point cloud stitching procedure instead of Voxblox++~\cite{grinvald2019volumetric}.
The bulk of the computation was carried out by the on-board CPU, with the exception of the object detection via YOLOv3, running on an on-board laptop-grade GPU, and the grasp planning and execution modules, offloaded to a third on-board CPU.
A video of the robot successfully performing the task is available online\footnote{\url{https://youtu.be/e89_Xg1sLnY}}.

\section{DISCUSSION AND OUTLOOK}

Our system is able to successfully navigate in indoor environments and manipulate objects. However, our approach has some limitations, providing insights into high-potential research directions.

The obstacle detection system is only able to detect obstacles in an ankle-height plane through the LiDARs on the mobile base. Being able to handle a wider range of obstacles, such as different types of tables, would be a big improvement to the reliability of the system. For example, some tables have legs which are not in the corners of the table. Our approach might cut through the corner causing a collision of the robot body with the table top. Measurements from an RGB-D sensor could be integrated into a 3D representation. This representation could then be used with the known robot mesh to plan paths that are collision free with the full robot.

The motion planner we used for navigation generates quite jerky trajectories for the base. Adding a trajectory optimization step could help us achieve smoother motion. The optimizer could minimize changes in acceleration in addition to avoiding obstacles.

Our current 3D object detection system is very basic and does not produce very reliable estimates of the object position. After the first 30-frame sequence of detections, we directly triangulate the object and place it in our map, regardless of the triangulation baseline. New observations of the object are not incorporated and no measure of uncertainty is associated with the detected object position. Maintaining (potentially) multiple tracks of the detected object and updating the position estimate using a Bayesian filter can improve detection robustness. We could also incorporate additional sensing modalities, such as the depth channel, to directly gauge the 3D position of the object.

We used simple heuristics to position the base of the robot next to the manipulation target. A more comprehensive system would be needed to be able to deal with the full variety of workstations out there. A 3D obstacle detection system combined with a full-body planner would enable operation across a broader range of environments.

At present, the scanning trajectory during the dense 3D reconstruction step consists of a set of manually predefined views. 
While effective for the task at hand, this could be replaced by an active scanning approach, potentially yielding more efficient exploration and higher scene coverage.

The grasp planner provides good grasp proposals and is mostly agnostic to the type of object being grasped. While the proposed grasps are reasonable, they do not consider whether they are feasible in practice regarding kinematic constraints, potential collisions, etc. 
A more comprehensive grasp planner would reason about the feasibility of grasps.
Our grasp execution is currently open loop, thus there is no feedback during execution. We expect future dexterous robots to use closed-loop manipulation planning and control algorithms. In this case, feedback could be used to either adjust or completely rethink the grasping strategy, for example, to enable recovery maneuvers or pre-grasp contact.

Currently, our system's actions are designed to either operate the arm or the base exclusively. Combining all degrees of freedom of the system using a whole-body motion planning and control framework has the potential to render current actions more efficient. For example, in case the target object is out of reach, the base position could be adjusted simultaneously to the grasp being executed, instead of in an alternating fashion. Further, whole-body control opens up new capabilities that cannot be achieved by arm motion alone, e.g. opening of doors with large radii. 

Finally, our system has limited capability to deal with unexpected failures. If any of the individual steps of the task fail, we are unable to recover. If the banana falls out of the hand while bringing it to the goal, this is not detected. A more flexible plan execution framework could enable reactivity. Since introducing reactivity using state machines renders the execution software complex and hard to maintain and to adapt to other tasks in the future, it would be worthwhile investigating alternatives such as behavior trees~\cite{colledanchise2018behavior}.

\addtolength{\textheight}{-12cm}   



\bibliographystyle{IEEEtranN}
\footnotesize
\bibliography{bibliography}

\begin{acronym}
\acro{imu}[IMU]{Inertial Measurement Unit}
\acro{dof}[DoF]{Degrees of Freedom}
\acro{tsdf}[TSDF]{Truncated Signed Distance Function}
\acro{gpd}[GPD]{Grasp Pose Detection}
\end{acronym}

\end{document}